# Robust Point Cloud Registration in Robotic Inspection with Locally Consistent Gaussian Mixture Model

Ling-jie Su, Wei Xu*, and Wen-long Li, *Member, IEEE*

*Abstract*—In robotic inspection of aviation parts, achieving accurate pairwise point cloud registration between scanned and model data is essential. However, noise and outliers generated in robotic scanned data can compromise registration accuracy. To mitigate this challenge, this article proposes a probability-based registration method utilizing Gaussian Mixture Model (GMM) with local consistency constraint. This method converts the registration problem into a model fitting one, constraining the similarity of posterior distributions between neighboring points to enhance correspondence robustness. We employ the Expectation Maximization algorithm iteratively to find optimal rotation matrix and translation vector while obtaining GMM parameters. Both E-step and M-step have closed-form solutions. Simulation and actual experiments confirm the method's effectiveness, reducing root mean square error by 20% despite the presence of noise and outliers. The proposed method excels in robustness and accuracy compared to existing methods.

## NOMENCLATURE

| | |
|---|---|
| $N$ | Number of points in the scanned point cloud. |
| $M$ | Number of points in the model point cloud. |
| $X = \{\boldsymbol{x}_n, n = 1, 2, \ldots, N\}$ | Scanned point cloud. |
| $Y = \{\boldsymbol{y}_m, m = 1, 2, \ldots, M\}$ | Model point cloud. |
| $\boldsymbol{R}$ | Rotation matrix applied to $Y$. |
| $\boldsymbol{t}$ | Translation vector applied to $Y$. |
| $\phi(\boldsymbol{y}_m)$ | Transformation of point $\boldsymbol{y}_m$. |
| $\{\sigma_m^2\}_{m=1}^M$ | Variances in GMM. |
| $\Theta = \left\{\boldsymbol{R}, \boldsymbol{t}, \{\sigma_m^2\}_{m=1}^M\right\}$ | Parameter set to be estimated. |
| $\mathcal{Z} = \{z_n, n = 1, 2, \ldots, N\}$ | Hidden variables. |
| $p_{mn}$ | Abbreviation of posterior. |
| $\pi_m$ | Prior probability $p(z_n = m)$. |
| $D(\bullet \| \bullet)$ | KL-Divergence between two probability distribution. |
| $\lambda$ | The weight of the local consistency term. |
| $(\bullet)^{\mathrm{T}}$ | Transpose of a matrix or a vector. |
| $\det(\bullet)$ | Determinant of a matrix. |
| $\mathrm{Tr}(\bullet)$ | Trace of a matrix. |
| $\mathrm{diag}(\bullet)$ | Diagonal matrix of a vector. |
| $\|\bullet\|$ | 2-norm of a vector. |
| $\|\bullet\|_{\mathrm{F}}$ | Frobenius norm of a matrix. |
| RMSE | Rooted mean squared error. |
| $e_R$ | Error of the rotational matrix. |
| $e_t$ | Error of the translation vector. |
| $\mathbf{I}$ | Identity matrix. |

## I. INTRODUCTION

IN aviation field, blade [1] and aircraft skin [2] are pivotal parts. Precise measurement of these parts is essential for machining, ensuring the safety, performance, and durability of aircraft structures. Optical three-dimensional (3D) scanners, valued for their high accuracy, non-contact nature, and adaptability, are extensively employed for part measurement, yielding point cloud data [3]-[5]. In practical scenarios, 3D sensors are routinely integrated at the terminus of industrial robots to extend measurement capabilities [6]-[8], thereby capturing comprehensive point cloud representations, which is called robotic inspection, as shown in Fig. 1. In the robotic inspection of aviation parts, one crucial task is the registration of the scanned point cloud data with the point cloud of the standard model to obtain key parameters of these parts [9]-[11], for instance, the profile accuracy of aviation blade at various blade heights, aircraft skin machining quality, and skin-hole positions. Therefore, to ensure a dependable assessment of key parameters of aviation parts, a registration approach in the robotic inspection boasting high accuracy and robustness, is indispensable. Nonetheless, scanning results from the robotic inspection procedure, particularly those deployed in aviation industrial settings, often yield noisy point cloud data due to sensor instability, robotic instability, reflective surface of metal parts, and complex machining site environments [12]. The registration accuracy could be impaired by the noise. Additionally, outliers [13], stemming from background objects like clamps, robot ontology, and pedestals, further undermine registration accuracy. Therefore, mitigating the adverse effects of noise and outliers in the robotic inspection of aviation parts is pivotal for enhancing registration precision.

This work has been submitted to the IEEE for possible publication. Copyright may be transferred without notice, after which this version may no longer be accessible.

This project was supported by the National Natural Science Foundation of China under Grant 52205524, Grant 52188102, and Grant 52075203. (*Corresponding author: Wei Xu.*)

Ling-jie Su, Wei Xu, and Wen-long Li are with the State Key Laboratory of Intelligent Manufacturing Equipment and Technology, Huazhong University of Science and Technology, Wuhan 430074, China (e-mail: ljsu@hust.edu.cn, weixu.chn@gmail.com, wlli@mail.hust.edu.cn).

Color versions of one or more of the figures in this article are available online at http://ieeexplore.ieee.org

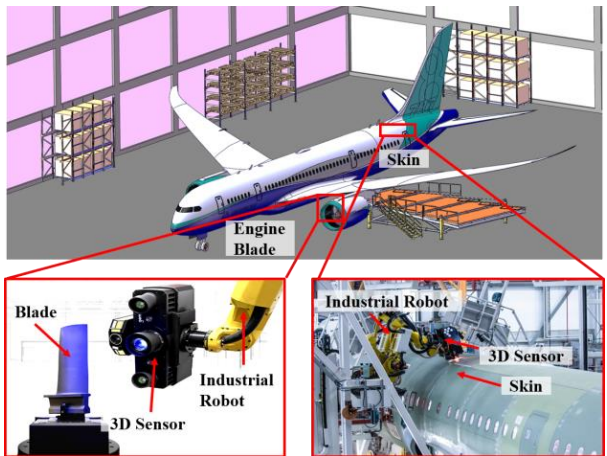

Fig. 1. 3D sensor equipped with an industrial robot in aviation part inspections.

### A. Related Work

Rigid pairwise registration methods can be roughly divided into three categories according to the strategy of searching correspondences between model point cloud and scanned point cloud, including distance-based, feature-based and probability-based methods. In this section, we review the development of registration methods in the three categories respectively.

The iterative closest point (ICP) registration method proposed by Besl et al. [14] is the classical distance-based method, which establishes the objective function minimizing the Euclidean distance between closest point correspondence in two point clouds. The key idea is to take the closest point as the substitute of the ground-truth correspondence. Many researchers have developed improved methods of ICP to enhance the performance, for instance, trimmed-ICP [15], Levenberg-Marquardt ICP (LM-ICP) [16], globally optimal algorithm (Go-ICP) [17], truncated least squares estimation and semidefinite relaxation (TEASER) algorithm [18]. These methods take point-to-point distance as the optimization object. Besides, some methods utilize the point-to-plane distance [19]-[21] to obtain better performance on data with complex surface. Segal et al. [22] proposed a generalized ICP method, designing arbitrary covariance matrices that include both point-to-point and point-to-plane. Distance-based registration methods have shown considerable success in aligning point clouds. However, challenges arise due to complex measurement environments of aviation blades and skin components, leading to inevitable noise and outliers in the measured data. These noisy points, often not on the part surface, can accumulate measurement errors, adversely affecting the registration process. Furthermore, outliers may be misconstrued as valid correspondences, resulting in incorrect correspondences and diminishing registration accuracy. Consequently, the effectiveness and applicability of distance-based methods in robotic inspection of aerospace parts may decrease.

The feature-based methods extract correspondences using local or global geometric information. These methods iteratively minimize the distance between feature correspondences to obtain the transformation. Belongie et al. [23] proposed a shape context feature descriptor to offer a globally discriminative characterization, which enables the estimation of transformation using regularized thin-plate splines. Rusu et al. [24] proposed fast point feature histograms (FPFH), a pose-invariant local feature type ideal for point correspondence search in point cloud registration. Yang et al. [25] presented a local feature statistics histogram (LFSH) descriptor, encoding statistical properties on local depth, point density, and normal angles for comprehensive local shape geometry description. Some studies employ neural networks for robust feature correspondence search. Zeng et al. [26] presented a 3D convolution neural network named 3D-Match, leveraging local volumetric regions and feature descriptors. Deng et al. [27] utilized a local point pair features (PPF) descriptor in PPFNet training, incorporating a novel N-tuple loss and architecture to enhance local feature distinction and global information integration. While feature-based methods are commonly employed for coarse registration to establish an initial transformation matrix, their accuracy is compromised by inaccurate feature points resulting from descriptor calculation errors. Additionally, these methods fail to address noise and outliers, leading to decreased correspondence correctness. Consequently, they prove unsuitable for high-accuracy registration, particularly in the context of robotic inspection of aviation metal parts.

In the probability-based methods, the point cloud registration problem is reframed as a probability model fitting task. Parameters of the model are determined via maximum likelihood estimation (MLE) theory, yielding simultaneous rigid transformation. As hidden variables are present, correspondence relationship becomes one-to-all rather than one-to-one . Myronenko et al. [28] proposed Coherent Point Drift (CPD) algorithm, treating the model point cloud as centroids in a GMM while the target cloud is seen as GMM samples. Expectation Maximization (EM) algorithm is utilized to achieve model fitting task. Jian et al. [29] introduced Gaussian Mixture Model registration (GMMReg) algorithm, which reformulates the registration as aligning GMM-based point cloud to minimize L2 distance between corresponding mixtures. Horaud et al. [30] utilized ECM algorithm replacing EM, enhancing performance due to interdependent registration and model parameters. Evangelidis et al. [31] proposed the Joint Registration of Multiple Point Clouds (JRMPC) algorithm, treating all point clouds as samples from an unknown GMM. Min et al. [32] proposed a hybrid mixture model (HMM) based algorithm, using Von-Mises-Fisher mixture model (FMM) for orientation and GMM for position to incorporate positional and orientational data together. Eckart et al. [33] substituted classical GMM with a tree-based variant, drastically improving efficiency while maintaining accuracy and generality. Yuan et al. [34] pioneered neural network utilization for probability mixture model fitting, framing registration as minimizing Kullback-Leibler Divergence (KL-Divergence) between probability models. The one-to-all correspondence strategy in probability-based methods enhances robustness compared to distance-based and feature-based methods. However, noise and outliers can negatively impact the posterior distribution, particularly in robotic inspection of aviation parts with complex measurement environments, leading to high uncertainty. This uncertainty may result in incorrect correspondences and compromised registration accuracy. Therefore, imposing additional constraints on the posterior distribution is essential for point cloud registration in aviation parts with robotic inspection. These constraints would bolster the method's robustness when dealing with data prone to high levels of noise and outliers.

### B. Contribution

This article proposes a robust pairwise registration method, addressing noise and outliers by framing the issue as a probability model fitting problem. To mitigate the negative impact from noise and outliers, the proposed registration method incorporating the local consistency with the classical GMM, constraining the similarity [35] of posteriors between neighboring points, enhancing the robustness of the method. Solving the model fitting problem under EM framework, the article offers closed-form solutions for parameters in each iteration. The proposed method has the following advantages.

**1) Enhanced Robustness.** The approach considers a one-to-all correspondence to prevent local optima issues common in one-to-one correspondence methods. It constrains the similarity of posterior distribution among neighboring points to mitigate noise and outliers' influence. Comparing to alternatives, this method excels particularly in the presence of noise and outliers.

**2) Improved Accuracy.** Unlike other probability-based methods, the proposed method accounts for local consistency. This additional constraint ensures the correctness of correspondence and guarantees the registration accuracy. Extensive experiments confirm the superior accuracy of the proposed method over others.

This paper is organized as follows. Section II introduces the registration workflow. Section III discusses the proposed registration method in detail, including the closed-form solution in the EM algorithm. Section IV conducts experiments to verify the performance of the proposed method. Finally, in Section V, the conclusion is presented.

## II. REGISTRATION WORKFLOW WITH LCGMM

In this section, the frame of the proposed registration method utilizing locally consistent GMM (LCGMM) is presented, as shown in Fig. 2. All the notations are described in the Nomenclature section.

In the proposed method, the centroid of each component in the GMM is fixed to the transformed point in model point cloud $Y$ and the points in scanned point cloud $X$ are regarded as the samples from the GMM. The transformed model point is denoted as $\phi(y_m) = Ry_m + t$. Thus, the centroids of the GMM are related to the registration relationship between $X$ and $Y$. According to MLE theory, the parameters $\Theta$ can be estimated by maximizing the observed log-likelihood function as follows,

$$L(\Theta \mid X) = \log P(X; \Theta). \tag{1}$$

Due to the unknown component in GMM from which the observed point $x_n$ is sampled, it is unfeasible to maximize (1) directly. To describe the probability of $x_n$ being sampled from each component in GMM, hidden variables $\mathcal{Z} = \{z_n, n = 1, 2, \ldots, N\}$ are introduced and $z_n = m$ means that the $n$-th point is sampled from the $m$-th Gaussian component. The sampled data and the correspondent hidden variables are called complete data. EM algorithm is used to solve the MLE problem with the existence of hidden variables. In the EM algorithm, the expectation of the observed complete-data log-likelihood is to be maximized, which is as follows,

$$Q_{\text{GMM}}(\Theta \mid X, \mathcal{Z}) = E_{\mathcal{Z}}\left[\log P(X, \mathcal{Z}; \Theta \mid X)\right]. \tag{2}$$

The maximized $Q_{\text{GMM}}(\Theta \mid X, \mathcal{Z})$ is the lower bound of the maximal $L(\Theta \mid X)$ and the optimal parameters can be estimated by maximizing expectation iteratively.

To improve the robustness of the registration method against noise and outliers, the local geometric information is considered. The similarity of posteriors between neighboring points in the scanned point cloud is incorporated into the EM algorithm, resulting in the posteriors not likely being affected by noise and outliers to guarantee the registration accuracy with the existence of noise and outliers. The local consistency is incorporated into classical GMM as a regulation term $Q_{\text{LC}}$ with a parameter $\lambda$ to adjust the weight.

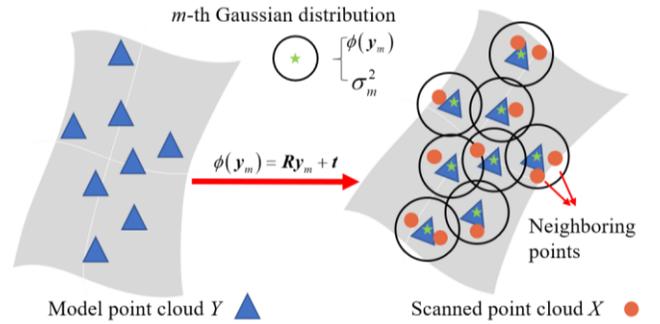

Fig. 2. Illustration of the proposed registration method. The transformed points in model point cloud $Y$ is regarded as the centroids of GMM and the points in the scanned point cloud $X$ are sampled from the GMM. The geometric information of neighboring points is incorporated into the GMM fitting procedure.

The workflow of the proposed method is as follows.

1) Initialize the parameters $\Theta$.

2) E-step. Calculate the posterior $p_{mn}^q$ with the current parameters, $q$ is the iteration index.

3) M-step1: Estimate the translation vector $t^q$ with the current posterior and GMM variances.

$$t^q = \arg\min_t \left\{ -E_{\mathcal{Z}}\left[\log P(X, \mathcal{Z}; \Theta_t \mid X)\right] + \lambda Q_{\text{LC}} \right\} \tag{3}$$

where $\Theta_t = \left\{ R, t, \{\sigma_m^2\}_{m=1}^M \right\}$.

4) M-step2: Estimate the rotation matrix $R^q$ with updated translation vector $t^q$

$$R^q = \arg\min_R \left\{ -E_{\mathcal{Z}}\left[\log P(X, \mathcal{Z}; \Theta_R \mid X)\right] + \lambda Q_{\text{LC}} \right\} \tag{4}$$

where $\Theta_R = \left\{ R, t^q, \{\sigma_m^2\}_{m=1}^M \right\}$.

5) M-step3: Estimate the variance $\sigma_m^2$ with updated translation vector $t^q$ and the rotation matrix $R^q$

$$\sigma_m^{2*} = \arg\min_{\sigma_m^2} \left\{ -E_{\mathcal{Z}}\left[\log P(X, \mathcal{Z}; \Theta_{\sigma^2} \mid X)\right] + \lambda Q_{\text{LC}} \right\} \tag{5}$$

where $\Theta_{\sigma^2} = \left\{ R^q, t^q, \{\sigma_m^2\}_{m=1}^M \right\}$.

6) Terminate the algorithm if the iteration number exceeds the threshold.

## III. METHOD

In this section, the LCGMM is used to conduct point cloud registration task. First, the optimization object based on probability model is established and then the EM algorithm is used to conduct the closed-form solution in each iteration. The rotation matrix and translation vector are obtained during the probability model fitting process simultaneously.

### A. Point cloud registration

Given the scanned point cloud $X = \{x_n, n = 1, 2, \ldots, N\}$ and the model point cloud $Y = \{y_m, m = 1, 2, \ldots, M\}$, $N$ and $M$ are the numbers of point clouds, the point cloud registration task is to find a rotation matrix $R$ and a translation vector $t$ such that the two point clouds can be aligned together. The transformed points in $Y$ can be seen as the centroid of each component in GMM, while the points in $X$ are observed samples from GMM. To describe the probability of $x_n$ being sampled from each Gaussian component, hidden variables $\mathcal{Z} = \{z_n, n = 1, 2, \ldots, N\}$ are defined. $z_n = m$ represents that $x_n$ is sampled from the $m$-th Gaussian component. The probability of the $n$-th point being sampled from the $m$-th Gaussian component is as follows,

$$p(x_n \mid z_n = m; \Theta) = \frac{1}{(2\pi\sigma_m^2)^{3/2}} e^{-\frac{1}{2\sigma_m^2}\|x_n - \phi(y_m)\|^2} \quad (6)$$

where $\Theta = \{R, t, \{\sigma_m^2\}_{m=1}^M\}$ are the parameters to be estimated. The probability of $x_n$ being observed is as follows,

$$p(x_n; \Theta) = \sum_{m=1}^{M} \pi_m p(x_n \mid z_n = m; \Theta) + \pi_{M+1} \mathcal{U}(V) \quad (7)$$

where $\mathcal{U}(V) = 1/V$ is the uniform distribution over the bounding volume $V$ of the scanned part and is indexed by $M+1$. $\pi_m$ is the prior information and $\pi_m = p(z_n = m)(0 < m < M+1)$ is a constant weight for each distribution given in advance. $\pi_{M+1}$ is the weight for outliers, set by a constant $\omega$ $(0 \leq \omega \leq 1)$ and $\sum_{m=1}^{M+1} \pi_m = 1$. To obtain the optimal parameters for GMM and transformation, the expectation of the negative complete-data log-likelihood is minimized iteratively, which is as follows,

$$Q_{\mathrm{GMM}}(\Theta \mid \Theta^{\mathrm{old}}) = -\sum_{\mathcal{Z}} P(\mathcal{Z} \mid X; \Theta^{\mathrm{old}}) \log P(X, \mathcal{Z}; \Theta). \quad (8)$$

$Q_{\mathrm{GMM}}(\Theta \mid \Theta^{\mathrm{old}})$ in (8) can be expanded as follows,

$$Q_{\mathrm{GMM}}(\Theta \mid \Theta^{\mathrm{old}}) = -\sum_{n=1}^{N} \sum_{m=1}^{M+1} p(z_n = m \mid x_n; \Theta^{\mathrm{old}}) \log(\pi_m p(x_n \mid z_n = m; \Theta)) \quad (9)$$

$$= \sum_{n=1}^{N} \sum_{m=1}^{M} p_{mn} \frac{\|x_n - \phi(y_m)\|^2}{2\sigma_m^2} + \frac{3}{2} \sum_{n=1}^{N} \sum_{m=1}^{M} p_{mn} \log \sigma_m^2$$

where the posterior $p(z_n = m \mid x_n; \Theta^{\mathrm{old}})$ is denoted as $p_{mn}$, which is constant in the M-step. The derivation from (8) to (9) can be found in Appendix A. In the registration procedure, the posterior $p_{mn}$ represents the probability of the $n$-th point being sampled from the $m$-th Gaussian component, which is the correspondent relationship between the $n$-th point in $X$ and the $m$-th point in $Y$. Thus, the correctness of $p_{mn}$ determines the accuracy of the registration result. To improve the robustness of the posterior against noise and outliers, the local geometric information in the scanned point cloud is used.

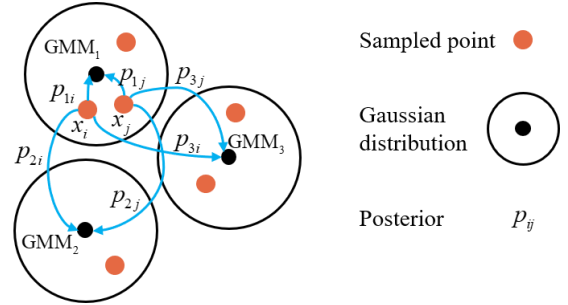

Fig. 3. Posteriors of two neighboring points $x_i$ and $x_j$.

According to [35], there is a local consistency propriety of the posterior in GMM, which is that $p(z_i \mid x_i)$ are similar within neighboring sampled points. The posteriors of two neighboring sampled points are shown in Fig. 3. It is intuitive that when two samples are close, the probabilities of them belonging to a certain component should be similar. From another aspect, neighboring points are more likely to belong to the same Gaussian component. The local consistency propriety strengthens posterior probability to the Gaussian component that the point is sampled from and reduces the influence of distant points. Integrating this property into the GMM fitting procedure mitigates the negative impact of noise and outliers, improving method's robustness and accuracy.

To integrate the local consistency propriety into the GMM fitting procedure, quantitatively weighing the similarity of posteriors between neighboring points is essential. KL-Divergence is a wildly-used merit to measure the similarity between two probability distribution, leveraging its non-negativity, asymmetry and relative properties. Given two posteriors $p(z_i \mid x_i)$ and $p(z_j \mid x_j)$, the KL-Divergence of $p(z_i \mid x_i)$ relating to $p(z_j \mid x_j)$ is as follows,

$$D\big(p(z_i \mid x_i) \parallel p(z_j \mid x_j)\big) = \sum_{m=1}^{M+1} p(z_i = m \mid x_i) \log \frac{p(z_i = m \mid x_i)}{p(z_j = m \mid x_j)}. \quad (10)$$

To make the similarity between the two distributions symmetric, the modified measurement is as follows,

$$D_{ij} = \frac{1}{2}\Big(D\big(p(z_i \mid x_i) \parallel p(z_j \mid x_j)\big) + D\big(p(z_j \mid x_j) \parallel p(z_i \mid x_i)\big)\Big). \quad (11)$$

The smaller of $D_{ij}$, the more similar the two posteriors are. The local consistency property can be added to the classical GMM as a regularization term as follows,

$$Q_{\text{LC}} = \sum_{i=1}^{N} \sum_{j=1}^{N} w_{ij} D_{ij} \tag{12}$$

where $w_{ij}$ represents whether $x_i$ and $x_j$ are close to each other. If $x_i$ and $x_j$ are neighbors to each other, $w_{ij} = 1$, otherwise $w_{ij} = 0$.

The new optimization object can be modified as minimizing the log-likelihood function with local consistency regularization term, which is as follows,

$$Q(\Theta) = Q_{\text{GMM}}\left(\Theta \mid \Theta^{\text{old}}\right) + \lambda Q_{\text{LC}} \tag{13}$$

where $\lambda$ is the regularization parameter, which can adjust the influence of the local consistency term and should be given in advance.

### B. Closed-form Solution with EM Algorithm

EM algorithm computes a closed-form solution to minimize Equation (13) in each iteration. During the E-step, the posteriors $p_{mn}$ are calculated and remain constant in the M-step. In the M-step, the registration parameters $R$, $t$, and the GMM parameters $\sigma_m^2$ are updated sequentially.

1) *E-step*: The posterior $p\left(z_n = m \mid x_n; \Theta^{\text{old}}\right)$ represents the probability of the observed point $x_n$ corresponding to the model centroid $y_m$. The Bayesian formula is used to obtain the posterior, which is as follows,

$$
\begin{aligned}
& p\left(z_n = m \mid x_n; \Theta^{\text{old}}\right) \\
&= \frac{\pi_m p\left(x_n \mid z_n = m; \Theta^{\text{old}}\right)}{p\left(x_n; \Theta^{\text{old}}\right)} \\
&= \frac{\pi_m \left(2\pi\sigma_m^2\right)^{-3/2} e^{-\frac{1}{2\sigma_m^2}\|x_n - \phi(y_m)\|^2}}{\sum_{m=1}^{M} \pi_m \left(2\pi\sigma_m^2\right)^{-3/2} e^{-\frac{1}{2\sigma_m^2}\|x_n - \phi(y_m)\|^2} + \pi_{M+1}\frac{1}{V}}.
\end{aligned} \tag{14}
$$

2) *Derivation of regularization term*: With the specific posterior, we can deduct the specific form of $Q_{\text{LC}}$. Combining (14) with (10), the similarity measurement between $p\left(z_i \mid x_i\right)$ and $p\left(z_j \mid x_j\right)$ can be denoted as follows,

$$
\begin{aligned}
D_{ij} = \sum_{m=1}^{M} \frac{1}{4\sigma_m^2}\left(p_{mi} - p_{mj}\right) \\
\times \left(\|x_j - \phi(y_m)\|^2 - \|x_i - \phi(y_m)\|^2\right)
\end{aligned} \tag{15}
$$

where $p_{mi}$ and $p_{mj}$ are the corresponding posteriors $p\left(z_i = m \mid x_i; \Theta^{\text{old}}\right)$ and $p\left(z_j = m \mid x_j; \Theta^{\text{old}}\right)$. The detailed derivation of (15) is shown in Appendix B. With the detailed form of expectation of the negative complete-data log-likelihood and the local consistency constraint, the detailed form of the objective function to be minimized is as follows,

$$
\begin{aligned}
& Q(\Theta) \\
&= \sum_{n=1}^{N} \sum_{m=1}^{M} p_{mn} \frac{\|x_n - \phi(y_m)\|^2}{2\sigma_m^2} + \frac{3}{2} \sum_{n=1}^{N} \sum_{m=1}^{M} p_{mn} \log \sigma_m^2 \\
&\quad + \lambda \sum_{i=1}^{N} \sum_{j=1}^{N} w_{ij} \sum_{m=1}^{M} \left[\frac{1}{4\sigma_m^2}\left(p_{mi} - p_{mj}\right)\right. \\
&\quad \left. \times \left(\|x_j - \phi(y_m)\|^2 - \|x_i - \phi(y_m)\|^2\right)\right].
\end{aligned} \tag{16}
$$

3) *M-step*: In this step, the parameters $t$, $R$, and $\sigma_m^2$ are updated successively. Ignoring the constant relating to the translation vector $t$, the objective function (16) can be rewritten as follows,

$$
\begin{aligned}
Q(t) = \sum_{n=1}^{N} \sum_{m=1}^{M} \frac{p_{mn}}{2\sigma_m^2}\left(t^{\text{T}}t + 2y_m^{\text{T}}R^{\text{T}}t - 2x_n^{\text{T}}t\right) \\
- \lambda \sum_{i=1}^{N} \sum_{j=1}^{N} w_{ij} \sum_{m=1}^{M} \frac{p_{mi} - p_{mj}}{2\sigma_m^2}\left(x_j - x_i\right)^{\text{T}} t.
\end{aligned} \tag{17}
$$

The optimal value of $t$ can be obtained by solving $\partial Q(t)/\partial t = 0$, the results are as follows,

$$t^* = \mu_x - R\mu_y$$

$$\mu_x = \frac{\sum_{n=1}^{N} \sum_{m=1}^{M} \frac{p_{mn}}{\sigma_m^2}x_n + \frac{\lambda}{2} \sum_{i=1}^{N} \sum_{j=1}^{N} w_{ij}\left(x_j - x_i\right) \sum_{m=1}^{M} \frac{p_{mi} - p_{mj}}{\sigma_m^2}}{\sum_{n=1}^{N} \sum_{m=1}^{M} \frac{p_{mn}}{\sigma_m^2}} \tag{18}$$

$$\mu_y = \frac{\sum_{n=1}^{N} \sum_{m=1}^{M} \frac{p_{mn}}{\sigma_m^2}y_m}{\sum_{n=1}^{N} \sum_{m=1}^{M} \frac{p_{mn}}{\sigma_m^2}}.$$

Similar to $t^*$, the optimal value of $R$ can be obtained by solving $\partial Q(\Theta)/\partial R = 0$ with two constraints: 1) $R^{\text{T}}R = I$, 2) $\det(R) = 1$. Ignoring the constants relating to $R$ and supplanting the translation vector by $t^*$, the objection function (16) can be rewritten as follows,

$$
\begin{aligned}
& Q(R) = \sum_{n=1}^{N} \sum_{m=1}^{M} p_{mn} \frac{\|x_n - \phi(y_m)\|^2}{2\sigma_m^2} \\
&\quad + \lambda \sum_{i=1}^{N} \sum_{j=1}^{N} w_{ij} \sum_{m=1}^{M} \left[\frac{1}{4\sigma_m^2}\left(p_{mi} - p_{mj}\right)\right. \\
&\quad \left. \times \left(\|x_j - \phi(y_m)\|^2 - \|x_i - \phi(y_m)\|^2\right)\right].
\end{aligned} \tag{19}
$$

By setting $x_n' = x_n - \mu_x$, $y_m' = y_m - \mu_y$ and ignoring constants, the optimal rotation matrix can be obtained by solving the function as follows,

$$
\begin{aligned}
R^* = \arg\max_{R} \left(\sum_{n=1}^{N} \sum_{m=1}^{M} \frac{p_{mn}}{\sigma_m^2} x_n'^{\text{T}}R y_m'\right. \\
\left. + \frac{\lambda}{2} \sum_{i=1}^{N} \sum_{j=1}^{N} w_{ij} \sum_{m=1}^{M} \frac{p_{mi} - p_{mj}}{\sigma_m^2}\left(x_j' - x_i'\right)^{\text{T}} R y_m'\right).
\end{aligned} \tag{20}
$$

By utilizing the property of matrix trace, (20) can be transformed to maximize the trace of the product of two matrices, which is as follows,

$$\boldsymbol{R}^* = \arg\max_{\boldsymbol{R}}\left(\mathrm{Tr}\left(\boldsymbol{R}\mathbf{H}\right)\right) = \arg\max_{\boldsymbol{R}}\left(\mathrm{Tr}\left(\boldsymbol{R}\left(\mathbf{H}_1 + \mathbf{H}_2\right)\right)\right)$$

$$\mathbf{H}_1 = \sum_{n=1}^{N}\sum_{m=1}^{M}\frac{p_{mn}}{\sigma_m^2}\boldsymbol{y}_m'\boldsymbol{x}_n'^{\mathrm{T}} \qquad (21)$$

$$\mathbf{H}_2 = \frac{\lambda}{2}\sum_{i=1}^{N}\sum_{j=1}^{N}w_{ij}\frac{p_{mi}-p_{mj}}{\sigma_m^2}\boldsymbol{y}_m'\left(\boldsymbol{x}_j'-\boldsymbol{x}_i'\right)^{\mathrm{T}}.$$

Performing the singular value decomposition on $\mathbf{H}$, $\mathbf{H} = \mathbf{U}\mathbf{S}\mathbf{V}^{\mathrm{T}}$. The optimal value of the rotation matrix is as follows,

$$\boldsymbol{R}^* = \mathbf{V}\operatorname{diag}\left(\left[1, 1, \det\left(\mathbf{V}\mathbf{U}^{\mathrm{T}}\right)\right]\right)\mathbf{U}^{\mathrm{T}}. \qquad (22)$$

The detailed derivations are shown in Appendix C.

The optimal values of $\left(\sigma_m^2\right)_{m=1}^{M}$ are calculated by solving the equation $\partial Q(\Theta)/\partial\sigma_m^2 = 0$, the objective function ignoring the constants relating to $\sigma_m^2$ is as follows,

$$
\begin{aligned}
Q\left(\sigma_m^2\right) &= \sum_{n=1}^{N}p_{mn}\frac{\left\|\boldsymbol{x}_n - \phi(\boldsymbol{y}_m)\right\|^2}{2\sigma_m^2} + \frac{3}{2}\sum_{n=1}^{N}p_{mn}\log\sigma_m^2 \\
&+ \lambda\sum_{i=1}^{N}\sum_{j=1}^{N}\left[w_{ij}\frac{1}{4\sigma_m^2}\left(p_{mi}-p_{mj}\right)\right. \\
&\left.\times\left(\left\|\boldsymbol{x}_j - \phi(\boldsymbol{y}_m)\right\|^2 - \left\|\boldsymbol{x}_i - \phi(\boldsymbol{y}_m)\right\|^2\right)\right].
\end{aligned}
\qquad (23)
$$

The value of $\sigma_m^{2*}$ is as follows,

$$
\sigma_m^{2*} = \frac{\displaystyle\sum_{n=1}^{N}p_{mn}\left\|\boldsymbol{x}_n - \phi(\boldsymbol{y}_m)\right\|^2}{3\displaystyle\sum_{n=1}^{N}p_{mn}}
$$
$$
+ \frac{\dfrac{\lambda}{2}\displaystyle\sum_{i=1}^{N}\sum_{j=1}^{N}w_{ij}\left(p_{mi}-p_{mj}\right)\left(\left\|\boldsymbol{x}_j - \phi(\boldsymbol{y}_m)\right\|^2 - \left\|\boldsymbol{x}_i - \phi(\boldsymbol{y}_m)\right\|^2\right)}{3\displaystyle\sum_{n=1}^{N}p_{mn}}.
\qquad (24)
$$

Algorithm 1 showcases the proposed registration method.

| **Algorithm 1:** Robust point cloud registration with LCGMM |
| --- |
| 1: **Initialization:** Initialize parameter set $\Theta^0$ |
| 2: current iteration number $q = 1$ |
| 3: **if** not converge |
| 4:     E-step: Compute posterior $p_{mn}^q$ using (14). |
|     M-step: |
| 5:     a). Update rotation matrix $\boldsymbol{t}^q$ using (18) |
| 6:     b). Update rotation matrix $\boldsymbol{R}^q$ using (22) |
| 7:     c). Update variances $(\sigma_m^2)^q$ using (24) |
| 8:     $q = q + 1$ |
| 9: **end** |
| 10: **return:** $\boldsymbol{R}^*,\ \boldsymbol{t}^*$ |

## IV. Experiments

To validate the proposed method's performance, this section conducts experiments on both simulated and real-scanned data. Three registration methods, namely ICP, ECMPR, and JRMPC, are compared with the proposed method. ICP, widely used in industrial scenarios, serves as a benchmark. ECMPR and JRMPC are two probability-based methods described in the state-of-the-art. The simulation experiment assesses the effectiveness of the local consistency constraint and the proposed. method's robustness to varying levels of noise and outliers. In the actual experiment, the real-scanned blade and skin sample data are utilized to evaluate the proposed method's performance in real industrial settings. Qualitative and quantitative analyses of simulation and actual experiments are presented.

### A. Simulation experiment

In this section, we assess the proposed method's performance using the simulated blade data obtained through a simulation scanning on a blade model in Geomagic software. We conduct three separate sub-experiments: 1) The effectiveness of local consistency constraint is evaluated to determine its impact on the registration result; 2) The method's robustness to outliers is assessed by introducing outliers with varying ratios; 3) The method's robustness to noise is assessed by introducing Gaussian noise with different standard deviations. Fig. 4 illustrates the blade model, while Fig. 5 depicts the simulated-scanned point cloud with added noise and outliers. The model point cloud $Y$ comprises 5,000 points. The scanned point cloud $X$ is sampled differently in each sub-experiment. Gaussian noise and outliers with different levels are added to the scanned point clouds.

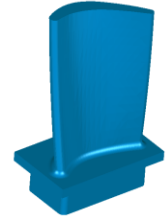

Fig. 4. Blade model in the simulation experiment.

Three metrics are utilized to quantitively evaluate the experiment results, which are root mean square error (RMSE), rotation error ($e_R$), and translation error ($e_t$). RMSE is defined as follows,

$$\mathrm{RMSE} = \frac{1}{M}\sqrt{\sum_{m=1}^{M}\left\|\phi_{\mathrm{gt}}(\boldsymbol{y}_m) - \phi_{\mathrm{cal}}(\boldsymbol{y}_m)\right\|^2} \qquad (25)$$

where $\phi_{\mathrm{gt}}(\boldsymbol{y}_m)$ and $\phi_{\mathrm{cal}}(\boldsymbol{y}_m)$ are the ground-truth and calculated transformation of the model point $\boldsymbol{y}_m$ separately. The rotation error $e_R$ is the Frobenius norm of the difference between the ground-truth and calculated rotation matrix, which is defined as follows,

$$e_R = \left\|\boldsymbol{R}_{\mathrm{gt}} - \boldsymbol{R}^*\right\|_{\mathrm{F}} \qquad (26)$$

where $\left\|\bullet\right\|_{\mathrm{F}}$ represents the Frobenius norm of a matrix, $\boldsymbol{R}_{\mathrm{gt}}$ and $\boldsymbol{R}^*$ are the ground-truth and the calculated rotation matrix respectively. The translation error $e_t$ is defined as the 2-norm of the difference between the ground-truth and calculated translation vector, which is defined as follows,

$$e_t = \left\| \boldsymbol{t}_{gt} - \boldsymbol{t}^* \right\| \tag{27}$$

where $\left\| \bullet \right\|$ represents the 2-norm of a vector. $\boldsymbol{t}_{gt}$ and $\boldsymbol{t}^*$ are the ground-truth and the calculated translation vector separately.

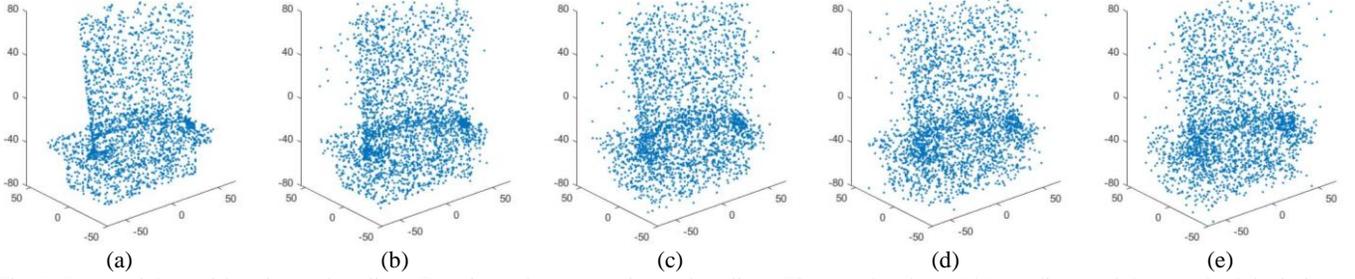

Fig. 5. Scanned data with noise and outliers. Data in (a) have no noise and outliers. The rest data have 10% outliers and the standard deviations of noise are 2.0, 3.0, 4.0, and 5.0 in (b)-(e) separately.

### 1) Experiment on the Effectiveness of Local Consistency

To test the effectiveness of local consistency constraint, the value of regulation term weight $\lambda$ is chosen differently in this experiment. We take the $\lambda$ value as 0.0, 0.4, 0.8, 1.2, 1.6, 2.0 respectively and conduct experiments on three scanned point clouds. The three scanned point clouds are sampled from the model point cloud $Y$ with 3,000, 4,000, and 5,000 points respectively, which are denoted as $X_1$, $X_2$, and $X_3$. Then, outliers with 10% ratio and noise with standard deviation valued as 4.0 are added to the scanned point clouds. For each trial, the ground-truth rotation matrix $\boldsymbol{R}_{gt}$ and the translation vector $\boldsymbol{t}_{gt}$ are generated randomly. The rotational angle along each axis is within the range $\left[ -60°, 60° \right]$, while the moving distance along each axis is in the range $\left[ -10, 10 \right]$.

The experiment results depicted in Fig. 6 indicate the influence of the local consistency constraint. When $\lambda$ is 0.0, the local consistency constraint does not influence the optimization process. It can be found that when $\lambda$ increases from 0.0, registration accuracy improves, enhancing the original GMM-based model's performance by this constraint. The local consistency term imposes constraints to the posterior similarity, improving one-to-all correspondences and enhancing registration robustness by mitigating noise and outliers. Beyond a weight value, further increases in $\lambda$ lead to diminishing improvements in registration errors due to excessive weights of the local consistency constraint, hindering the impact of expectation of complete-data log-likelihood. This experiment confirms the effectiveness of local consistency constraint.

### 2) Experiment on Robustness to Outliers

To test the proposed method's resilience to outliers, we add outliers at varying ratios ( 5% - 40% with a 5% interval) to the scanned point clouds $X_1$, $X_2$, and $X_3$. The standard deviation of Gaussian noise remains fixed at 4.0 in this sub-experiment.

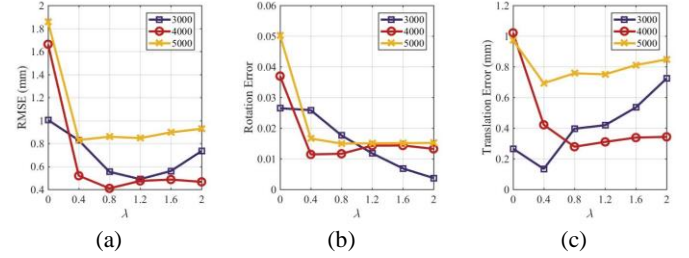

Fig. 6. The registration errors in the experiment on the effectiveness of local consistency. (a) RMSE. (b) $e_R$. (c) $e_t$.

The misalignments between the scanned point cloud and the model point cloud replicate those of the previous experiment. The regulation term weight $\lambda$ is set to be 0.5. The results depicted in Fig. 7 reveal a trend across three scanned point clouds: errors increase as the outlier ratio rises.

This correlation is expected, as outliers hinder correspondence searching. Among the four algorithms, ICP performs poorly, particularly with high outlier ratios. This is due to outliers being treated equally to inliers in the one-to-one correspondence, diminishing the impact of correct correspondences. The rest three algorithms utilize the one-to-all correspondence strategy outperforming ICP in accuracy. However, ECMPR and JRMPC solely use original GMM without considering the local consistency, leading to susceptibility to noise and outliers, thus reducing registration accuracy. Comparatively, the proposed method accounts for posterior similarity, mitigating invalid correspondences with outliers and enhancing the correspondence robustness. In most trials, RMSE, $e_R$, and the proposed method have the smallest value. Though the $e_R$ value of the proposed method and JRMPC are close in some trials, as shown in Fig. 7, the RMSE of the proposed method is still lower than that of JRMPC owing to the low value of $e_t$. The experiment results affirm the proposed method's robustness against outliers.

### 3) Experiment on Robustness to Noise

To verify the robustness of the proposed method to noise at different levels, we conduct tests using point cloud data with manually added Gaussian noise of varying standard deviations ($\sigma$ =2.0, 3.0, 4.0, 5.0). The misalignments remain consistent across the initial experiment, with a regulation term weight $\lambda$ set at 0.5 and a fixed outlier ratio of 10% . The numbers of the scanned point clouds are chosen to be 3,000, 4,000, and 5,000.

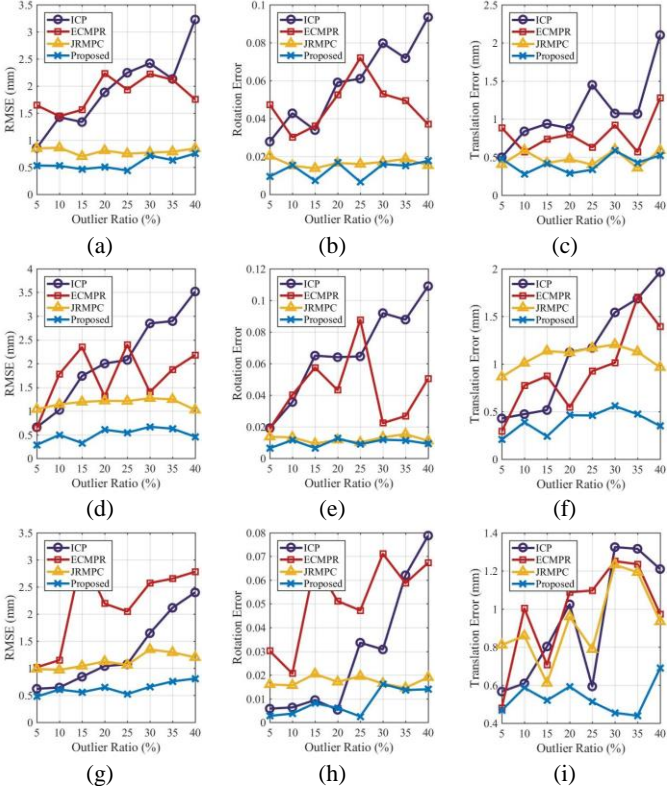

Fig. 7. The registration errors in the experiment on robustness to outliers. (a), (b), and (c) are RMSE, $e_R$, and $e_t$ for $X_1$. (d), (e), and (f) are RMSE, $e_R$, and $e_t$ for $X_2$. (g), (h), and (i) are RMSE, $e_R$, and $e_t$ for $X_3$.

To statistically compare the performance of four methods, six trials are repeated under the same number of scanned point clouds and the standard derivation of Gaussian noise. Each trial involves resampling the scanned point cloud from the model point cloud and regenerating misalignments. The means and standard deviations of registration error are presented in Fig. 8. The comparison of RMSE results reveals that ICP and ECMPR underperform relative to other methods under low noise standard deviation, while the proposed method slightly outperforms JRMPC. Conversely, at high noise standard deviation, JRMPC's performance sharply declines, contrasting with the stable performance of the proposed method. This stability arises from the local consistency constraint, mitigating noise's negative impact and enhancing correspondence accuracy. ICP's susceptibility to noise due to its one-to-one correspondence strategy and ECMPR's classical GMM representation lacking constraints on the posterior distribution make them vulnerable to noise-induced deviations. Meanwhile, JRMPC's optimization of GMM centers leads to heightened sensitivity to noise due to increased degrees of freedom. Consequently, this experiment highlights the proposed method's robustness across varying noise levels.

## B. Actual Experiment

In this experiment, the blade and skin sample are scanned using 3D scanner mounted on the industrial robot's end-effector. The coordinates under different positions are aligned by the marked points arranged around or on the measured parts in advance. The measurement scene of the blade and the skin sample are shown in Fig. 9 and Fig. 10 (a) separately. The blade data are acquired by PowerScan E equipped with ABB IRB 1,600 and the skin sample data are acquired by PowerScan-Auto 2.3M equipped with ABB IRB 1,600. Both blade and skin sample data are down-sampled. The blade data have 13,000 points and the skin sample data have 7,000 points. The model blade data are generated by simulation scanning in Geomagic software with 30,000 points. Due to the absence of the skin sample model, the skin sample data measured by ZEISS GOM ATOS SCANBOX 5,108, renowned for high accuracy, substitutes the model skin sample data. The supplemented data are down-sampled to 10,000 points, as depicted in Fig. 10 (b). Fig. 11 illustrates both model and scanned data of blade and skin sample. The blue points represent the model point cloud and the orange point represent the scanned point cloud. To further evaluate the performance of the proposed method, we down-sampled the points around leading edge, low-curvature area and trailing edge of the blade data to simulate the phenomenon of low data density in certain area, denoting as blade-I, blade-II, and blade-III, as shown in Fig. 12. Subsequently, the registration experiment on blade data is reconducted. Similarly, we use ICP, ECMPR, JRMPC, and the proposed method to align the model and scanned point cloud. Quantitative assessment of the registration results employs RMSE within a 10 mm distance threshold between the model and scanned data.

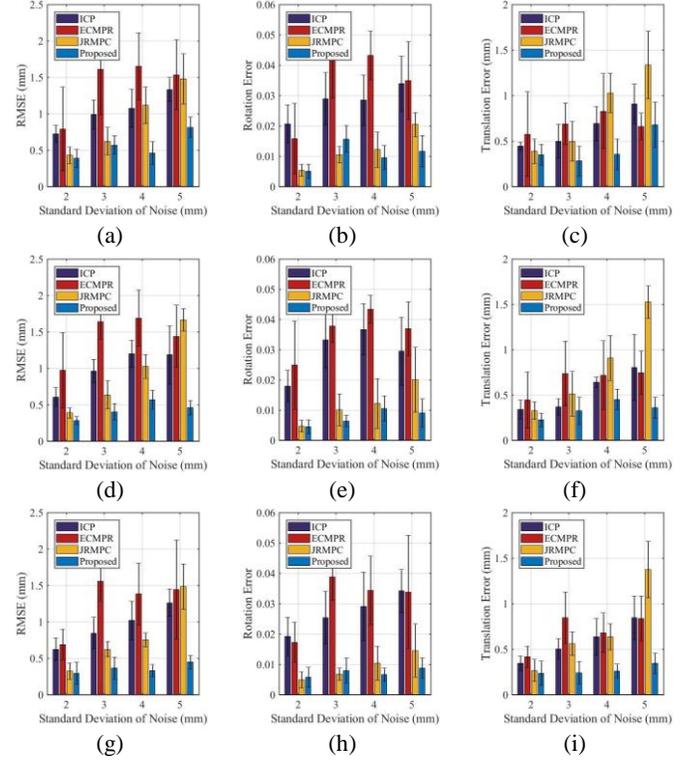

Fig. 8. The registration errors in the experiment on robustness to noise. (a), (b), and (c) are RMSE, $e_R$, and $e_t$ for sampled data with 3,000 points. (d), (e) and, (f) are RMSE, $e_R$, and $e_t$ for sampled data with 4,000 points. (g), (h), and (i) are RMSE, $e_R$, and $e_t$ for sampled data with 5,000 points.

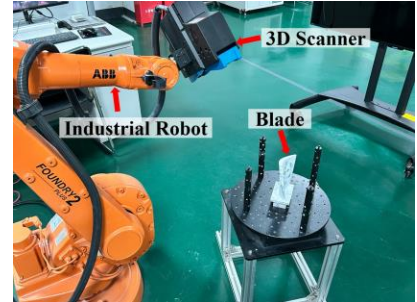

Fig. 9. The robotic inspection process of the blade.

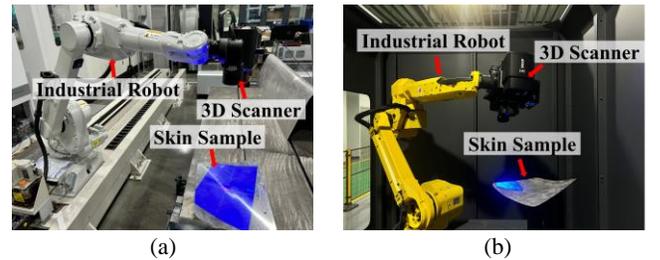

Fig. 10. (a) The robotic inspection process of the skin sample with PowerScan-Auto 2.3M sensor. (b) The measurement process of the skin sample with ZEISS GOM ATOS SCANBOX 5108.

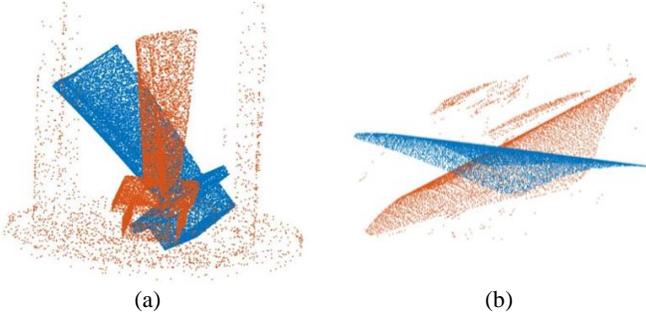

Fig. 11. Point cloud from robotic inspection of (a) model and scanned data of blade. (b) model and scanned data of skin samples.

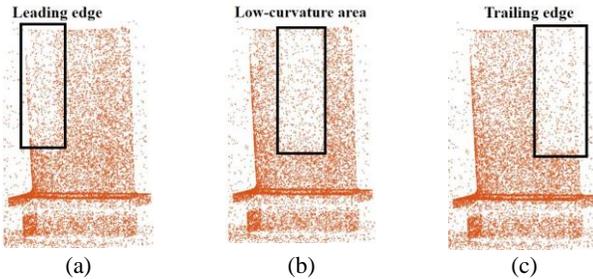

Fig. 12. Blade with low density points. (a) blade-I: low density in leading edge. (b) blade-II: low density in low-curvature area. (c) blade-III: low density in trailing edge.

The RMSE results of aligned blade, skin sample, blade-I, blade-II, and blade-III data are summarized in Table I, with the best highlighted in bold font. Comparative analysis reveals the superior performance of the proposed method, achieving complete overlap between the scanned and model point clouds, as shown in Fig. 13. Detailed registration information is provided through sections displaying aligned blade and skin sample data, illustrated in Fig. 14. Quantitative verification of section error is conducted by calculating the RMSE between model data sections and scanned data sections, as outlined in Table II. Sections of the scanned and model point cloud aligned by the proposed method demonstrate strong and the best overlap, highlighting the method's efficacy in detailed areas. This experiment shows the performance of the proposed method using real-scanned data. ICP's one-to-one correspondence strategy makes it vulnerable to outliers and the noise during robotic inspection. ECMPR lacks constraints on its posterior distribution, allowing noise and outliers to adversely affect correspondences. JRMPC, with centroids of GMM as parameters, has higher degrees of freedom , leading to susceptibility to noise and outliers. Comparatively, the proposed method is more robust to noise and outliers. The proposed method introduces a local consistency constraint to the posterior to enhance robustness, ensuring accurate correspondence relationship. Thus, it handles data with noise and outliers more effectively. The actual experiment in robotic inspection of aviation blade and skin parts validate the method's robustness and accuracy with real-scanned data in actual environments with noise and outliers.

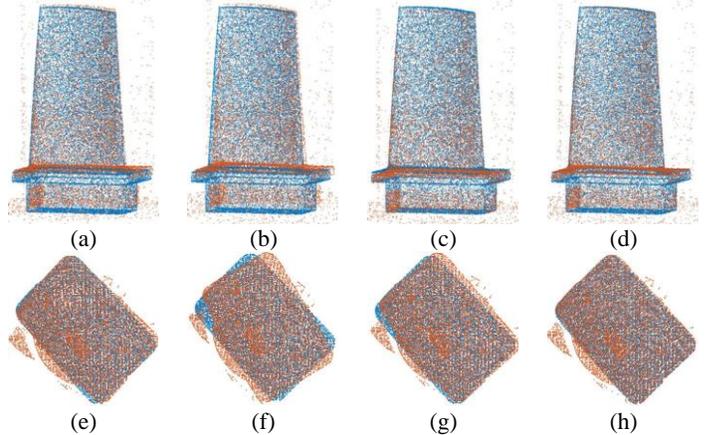

Fig. 13. Registration result of blade and skin sample data. (a)-(d) are results of blade data using ICP, ECMPR, JRMPC and the proposed method. (e)-(h) are the result of skin sample data using ICP, ECMPR, JRMPC and the proposed method.

### TABLE I
#### RMSE OF EXPERIMENT RESULTS

|           | ICP  | ECMPR | JRMPC | Proposed |
|-----------|------|-------|-------|----------|
| Blade     | 1.55 | 2.07  | 1.20  | **0.94** |
| Skin      | 3.07 | 3.90  | 2.21  | **2.01** |
| Blade-I   | 1.90 | 2.05  | 1.23  | **0.92** |
| Blade-II  | 1.99 | 2.26  | 1.26  | **0.86** |
| Blade-III | 2.15 | 2.62  | 1.24  | **0.98** |

### TABLE II
#### RMSE OF SECTIONS

|       | ICP  | ECMPR | JRMPC | Proposed |
|-------|------|-------|-------|----------|
| Blade | 0.76 | 1.04  | 0.46  | **0.36** |
| Skin  | 1.97 | 4.01  | 1.05  | **0.45** |

## V. CONCLUSION

To register the model point cloud with scanned point cloud with noise and outliers in robotic inspection of aviation parts, this article proposes a robust probability-based pairwise registration method. This method transforms the registration challenge into a GMM fitting problem. It integrates local consistency constraint of GMM into the optimization procedure to enhance correspondence's robustness and ensure the registration accuracy. EM algorithm optimizes the LCGMM model, iteratively providing closed-form solution. Simulation and actual experiments validate the proposed method's robustness and the accuracy. Experimental results demonstrate the effectiveness of the added local consistency constraint, outperforming three existing methods across varying levels of noise and outliers. Furthermore, actual experimental results in robotic inspection of two aviation parts showcase the proposed method's applicability and superiority of robustness and accuracy in actual industrial scenarios compared to the existing method.

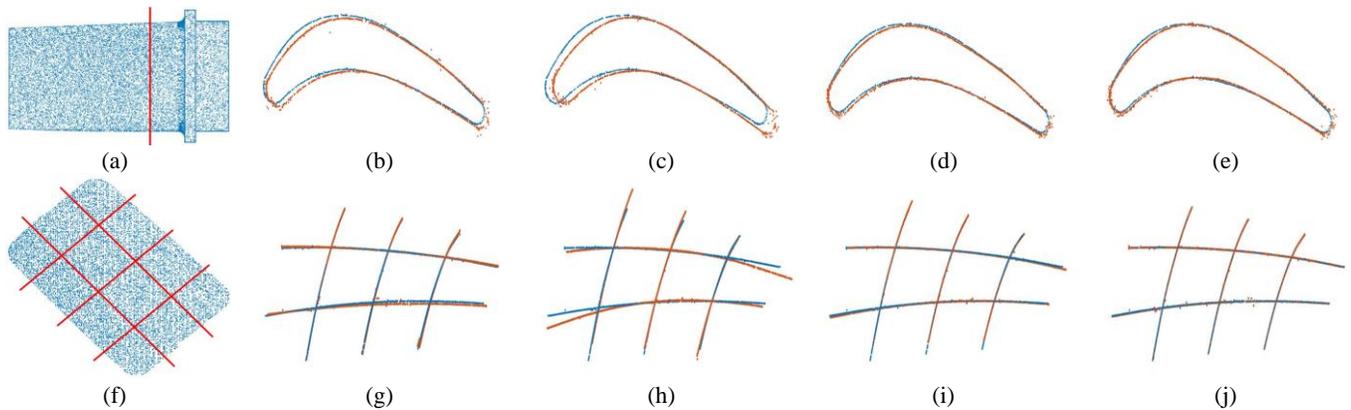

Fig. 14. (a) Blade section illustration. (b)-(e) are the blade section of registration results using ICP, ECMPR, JRMPC, and the proposed method. (f) skin sample section illustration. (g)-(j) are the skin sample section of registration results using ICP, ECMPR, JRMPC, and the proposed method.

## APPENDIX A
### DERIVATION OF (9)

For simplicity, the posterior $p\left(z_n = m \mid \boldsymbol{x}_n; \Theta^{\mathrm{old}}\right)$ is denoted as $p_{mn}$ in the M-step since they are constants. According to [32], (8) can be rewritten as,

$$Q\left(\Theta \mid \Theta^{\mathrm{old}}\right) = -\sum_{n=1}^{N}\sum_{m=1}^{M+1} p_{mn} \log\left(\pi_m p\left(\boldsymbol{x}_n \mid z_n = m; \Theta\right)\right). \quad (28)$$

Considering that $\pi_{M+1}$ and $p\left(\boldsymbol{x}_n \mid z_n = M+1; \Theta\right) = \mathcal{U}\left(\mathrm{V}\right)$ is constant in M-step, we have,

$$Q\left(\Theta \mid \Theta^{\mathrm{old}}\right) = -\sum_{n=1}^{N}\sum_{m=1}^{M} p_{mn} \log\left(\pi_m p\left(\boldsymbol{x}_n \mid z_n = m; \Theta\right)\right). \quad (29)$$

By substituting the expression of $p\left(\boldsymbol{x}_n \mid z_n = m; \Theta\right)$ in (6), the expression of $Q\left(\Theta \mid \Theta^{\mathrm{old}}\right)$ in (9) can be conducted.

## APPENDIX B
### DERIVATION $D_{ij}$ OF (15)

Combine the posterior in (14) with (10), we have,

$$D\left(p\left(z_i \mid x_i\right) \parallel p\left(z_j \mid x_j\right)\right)$$
$$= \sum_{m=1}^{M+1} p\left(z_i = m \mid x_i\right) \log \frac{p\left(z_i = m \mid x_i\right)}{p\left(z_j = m \mid x_j\right)}$$
$$= \sum_{m=1}^{M} p\left(z_i = m \mid x_i\right) \log \frac{p\left(z_i = m \mid x_i\right)}{p\left(z_j = m \mid x_j\right)}$$
$$\quad + p\left(z_i = M+1 \mid x_i\right) \log \frac{p\left(z_i = M+1 \mid x_i\right)}{p\left(z_j = M+1 \mid x_j\right)}. \quad (30)$$

By substituting $p\left(z_n = M+1 \mid x_n\right) = 1 - \sum_{m=1}^{M} p\left(z_n = m \mid x_n\right)$ into (30), we can get the following formula,

$$D\left(p\left(z_i \mid x_i\right) \parallel p\left(z_j \mid x_j\right)\right)$$
$$= \sum_{m=1}^{M} p_{mi} \frac{1}{2\sigma_m^2}\left(\left\|\boldsymbol{x}_j - \phi\left(\boldsymbol{y}_m\right)\right\|^2 - \left\|\boldsymbol{x}_i - \phi\left(\boldsymbol{y}_m\right)\right\|^2\right) \quad (31)$$
$$\quad + O\left(x_i \parallel x_j\right)$$

where,

$$O\left(x_i \parallel x_j\right)$$
$$= \log \frac{\sum_{m=1}^{M} \pi_m \left(2\pi\sigma_m^2\right)^{-\frac{3}{2}} e^{-\frac{1}{2\sigma_m^2}\left\|x_j - \phi\left(y_m\right)\right\|^2} + \pi_{M+1}\frac{1}{\mathrm{V}}}{\sum_{m=1}^{M} \pi_m \left(2\pi\sigma_m^2\right)^{-\frac{3}{2}} e^{-\frac{1}{2\sigma_m^2}\left\|x_i - \phi\left(y_m\right)\right\|^2} + \pi_{M+1}\frac{1}{\mathrm{V}}}. \quad (32)$$

Noting that $O(x_i \parallel x_j) + O(x_j \parallel x_i) = 0$, this term does not influence to $D_{ij}$. Thus, $O(x_i \parallel x_j)$ can be ignored. Equation (32) can be rewritten as follows,

$$D\left(p\left(z_i \mid x_i\right) \parallel p\left(z_j \mid x_j\right)\right)$$
$$= \sum_{m=1}^{M} p_{mi} \frac{1}{2\sigma_m^2}\left(\left\|\boldsymbol{x}_j - \phi\left(\boldsymbol{y}_m\right)\right\|^2 - \left\|\boldsymbol{x}_i - \phi\left(\boldsymbol{y}_m\right)\right\|^2\right). \quad (33)$$

Thus, $D_{ij}$ can be obtained by substituting (33) into (11).

## APPENDIX C
### DERIVATION $\boldsymbol{R}^*$ OF (19)

From (20), we have,

$$\boldsymbol{R}^* = \arg\max_{\boldsymbol{R}} \left( Q_1(\boldsymbol{R}) + Q_2(\boldsymbol{R}) \right) \qquad (34)$$

where,

$$Q_1(\boldsymbol{R}) = \sum_{n=1}^{N} \sum_{m=1}^{M} \frac{p_{mn}}{\sigma_m^2} \, {x'_n}^{\mathrm{T}} \boldsymbol{R} y'_m$$

$$Q_2(\boldsymbol{R}) = \frac{\lambda}{2} \sum_{i=1}^{N} \sum_{j=1}^{N} w_{ij} \sum_{m=1}^{M} \frac{p_{mi} - p_{mj}}{\sigma_m^2} \left( x'_j - x'_i \right)^{\mathrm{T}} \boldsymbol{R} y'_m. \qquad (35)$$

By utilizing the trace property $\mathrm{Tr}(\mathbf{ABC}) = \mathrm{Tr}(\mathbf{CAB})$ and $\mathrm{Tr}(\mathbf{A}+\mathbf{B}) = \mathrm{Tr}(\mathbf{A}) + \mathrm{Tr}(\mathbf{B})$, $Q_1(\boldsymbol{R})$ can be derived as,

$$
\begin{aligned}
Q_1(\boldsymbol{R}) &= \sum_{n=1}^{N} \sum_{m=1}^{M} \frac{p_{mn}}{\sigma_m^2} \mathrm{Tr}\left( {x'_n}^{\mathrm{T}} \boldsymbol{R} y'_m \right) \\
&= \sum_{n=1}^{N} \sum_{m=1}^{M} \frac{p_{mn}}{\sigma_m^2} \mathrm{Tr}\left( \boldsymbol{R} y'_m {x'_n}^{\mathrm{T}} \right) \\
&= \mathrm{Tr}\left( \sum_{n=1}^{N} \sum_{m=1}^{M} \frac{p_{mn}}{\sigma_m^2} \boldsymbol{R} y'_m {x'_n}^{\mathrm{T}} \right) \\
&= \mathrm{Tr}\left( \boldsymbol{R} \underbrace{\sum_{n=1}^{N} \sum_{m=1}^{M} \frac{p_{mn}}{\sigma_m^2} y'_m {x'_n}^{\mathrm{T}}}_{\mathbf{H}_1} \right).
\end{aligned}
\qquad (36)
$$

$Q_2(\boldsymbol{R})$ can be derived as,

$$
\begin{aligned}
Q_2(\boldsymbol{R}) &= \frac{\lambda}{2} \sum_{i=1}^{N} \sum_{j=1}^{N} w_{ij} \sum_{m=1}^{M} \frac{p_{mi} - p_{mj}}{\sigma_m^2} \mathrm{Tr}\left( \left( x'_j - x'_i \right)^{\mathrm{T}} \boldsymbol{R} y'_m \right) \\
&= \frac{\lambda}{2} \sum_{i=1}^{N} \sum_{j=1}^{N} w_{ij} \sum_{m=1}^{M} \frac{p_{mi} - p_{mj}}{\sigma_m^2} \mathrm{Tr}\left( \boldsymbol{R} y'_m \left( x'_j - x'_i \right)^{\mathrm{T}} \right) \\
&= \mathrm{Tr}\left( \frac{\lambda}{2} \sum_{i=1}^{N} \sum_{j=1}^{N} w_{ij} \sum_{m=1}^{M} \frac{p_{mi} - p_{mj}}{\sigma_m^2} \boldsymbol{R} y'_m \left( x'_j - x'_i \right)^{\mathrm{T}} \right) \\
&= \mathrm{Tr}\left( \boldsymbol{R} \underbrace{\frac{\lambda}{2} \sum_{i=1}^{N} \sum_{j=1}^{N} w_{ij} \sum_{m=1}^{M} \frac{p_{mi} - p_{mj}}{\sigma_m^2} y'_m \left( x'_j - x'_i \right)^{\mathrm{T}}}_{\mathbf{H}_2} \right).
\end{aligned}
\qquad (37)
$$

Thus, (34) can be written as,

$$
\begin{aligned}
\boldsymbol{R}^* &= \arg\max_{\boldsymbol{R}} \left( \mathrm{Tr}(\boldsymbol{R}\mathbf{H}_1) + \mathrm{Tr}(\boldsymbol{R}\mathbf{H}_2) \right) \\
&= \arg\max_{\boldsymbol{R}} \left( \mathrm{Tr}\left( \boldsymbol{R}(\mathbf{H}_1 + \mathbf{H}_2) \right) \right) \\
&= \arg\max_{\boldsymbol{R}} \left( \mathrm{Tr}(\boldsymbol{R}\mathbf{H}) \right)
\end{aligned}
\qquad (38)
$$

where $\mathbf{H} = \mathbf{H}_1 + \mathbf{H}_2$. According to [36], the orthonormal matrix $\mathbf{X} = \mathbf{V}\mathbf{U}^{\mathrm{T}}$ maximize $\mathrm{Tr}(\mathbf{XH})$, where $\mathbf{H} = \mathbf{U}\mathbf{S}\mathbf{V}^{\mathrm{T}}$. To make $\mathbf{X}$ satisfies the constraints of rotation matrix, we have,

$$\boldsymbol{R}^* = \mathbf{V} \mathrm{diag}\left( \begin{bmatrix} 1 & 1 & \det(\mathbf{V}\mathbf{U}^{\mathrm{T}}) \end{bmatrix} \right) \mathbf{U}^{\mathrm{T}}. \qquad (39)$$

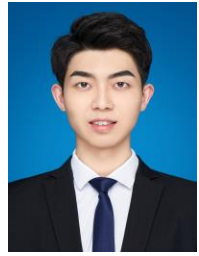

**Ling-jie Su** received the B.E. degree in Mechanical Design, Manufacturing and Automation from Huazhong University of Science & Technology, Wuhan, China, in 2022. He is currently working toward the master degree in mechanical engineering with Huazhong University of Science and Technology. His research interests are point cloud processing and 3D measurement.

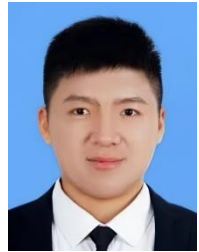

**Wei Xu** received the B.E. and M. Eng. degrees in fluid machinery and engineering from the China University of Mining & Technolog, China, in 2013 and 2016, respectively, and the Ph.D. degree in measurement engineering from Leibniz University Hannover, Hannover, Germany, in 2021. He is currently carrying out the postdoctoral research in mechanical engineering with the Huazhong University of Science & Technology, Wuhan, China. His research interests are the application of vision measurement and deep learning in robotic machining.

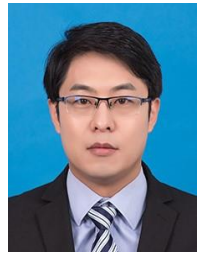

**Wen-long Li** (Member, IEEE) received the B.E. degree in Mechanical Engineering and Automation from Xi'an Jiaotong University, Xi'an, China, in 2004, the Ph.D. degree in Mechatronic Engineering from Huazhong University of Science & Technology (HUST), Wuhan, China, in 2010. Now he is a Professor at Huazhong University of Science & Technology, Wuhan, China. His current research interests include robotic machining and 3D optical measurement.